\documentclass[11pt]{article}
\usepackage{amsmath,amssymb,amsthm}
\usepackage[margin=1in]{geometry}
\usepackage{enumitem,url}
\usepackage{float}
\IfFileExists{algorithm.sty}{\usepackage{algorithm}}{%
  \floatstyle{ruled}\newfloat{algorithm}{tbp}{loa}\floatname{algorithm}{Algorithm}}
\usepackage[hidelinks]{hyperref}
\newcommand{\routine}[1]{\textsc{#1}}
\newenvironment{pseudocode}{\begin{enumerate}[label=\arabic*.,leftmargin=2em,itemsep=2pt,topsep=3pt]}{\end{enumerate}}
\newtheorem{theorem}{Theorem}
\newtheorem{lemma}[theorem]{Lemma}
\newtheorem{claim}{Claim}[theorem]
\theoremstyle{definition}

\newcommand{\E}{\mathbb E}

\newcommand{\R}{\mathbb R}
\newcommand{\ind}{\mathbf 1}
\newcommand{\cX}{\mathcal X}
\newcommand{\cH}{\mathcal H}

\newcommand{\Reg}{\mathrm{Reg}}
\newcommand{\KL}{\mathrm{KL}}

\newcommand{\Lap}{\mathrm{Lap}}
\newcommand{\Pois}{\mathrm{Pois}}

\newcommand{\argmax}{\operatorname*{arg\,max}}
\newcommand{\argmin}{\operatorname*{arg\,min}}
\newcommand{\norm}[1]{\lVert #1\rVert}

\setlist[enumerate]{topsep=4pt,itemsep=3pt,leftmargin=*}
\allowdisplaybreaks[2]
\title{Near-Optimal Regret for Oracle-Efficient\\
Adversarial Contextual Bandits}
\author{Kiarash Banihashem\\
University of Maryland, College Park\\
\texttt{kiarash@umd.edu}
\and
MohammadTaghi Hajiaghayi\\
University of Maryland, College Park\\
\texttt{hajiagha@umd.edu}}
\date{}
\begin{document}
\maketitle

\begin{abstract}
We study contextual bandits with $K$ actions, a class of $N$ policies,
and i.i.d.\ contexts from a distribution that the learner can sample.
For any sequence of loss functions fixed before play, we give an
oracle-efficient algorithm with expected regret
$O(\sqrt{KT\log(2N)}\,\log(2KT))$ against the best policy in hindsight
on the realized contexts. This improves the $T^{2/3}$ dependence on the horizon in prior
oracle-efficient bounds and achieves the optimal regret rate in this
setting up to logarithmic factors.

Our algorithm perturbs policy scores using a Poisson sample of auxiliary
contexts. At each sampled context, the perturbation assigned to an
action is a Laplace noise draw multiplied by a scale specific to that
context and action. We show that choosing these scales to minimize a
convex objective determined by past observations yields the stated regret bound. We further approximate
the minimizing scales by projected stochastic gradient descent using
only an exact offline cost-sensitive optimization oracle that returns
the optimum value. Since the table of scales can be large, the
algorithm represents the SGD iterates implicitly by storing the random
draws that determine each update and evaluates the scales only at
queried contexts. Both the oracle complexity and the remaining resource
bounds are polynomial in $T$, $K$, and $\log(2N)$.
\end{abstract}

\section{Introduction}

In a contextual bandit problem, the learner observes a context on each
of $T$ rounds, chooses one of $K$ actions, and observes only the loss of
that action. Regret is measured against the best policy in hindsight
from a class of $N$ policies. The classical \textsc{Exp4} algorithm
achieves expected regret $O(\sqrt{KT\log N})$ by maintaining a weight
for each policy~\cite{auer}. When the policy class is large, a natural
question is whether one can achieve this rate using only an offline
optimization oracle for the class. When contexts and loss vectors are
drawn jointly i.i.d., Langford and Zhang~\cite{epoch} obtained a regret
bound with $T^{2/3}$ dependence on the horizon, and Agarwal
et al.~\cite{agarwal} achieved the optimal square-root rate using an
offline optimization oracle.

Achieving square-root regret with an offline optimization oracle becomes
much more challenging however when losses are adversarial.
In this setting, Rakhlin and Sridharan~\cite{bistro} used
auxiliary context samples to obtain expected regret
$O(T^{3/4}K^{1/2}(\log N)^{1/4})$. Syrgkanis et al.~\cite{slks}
improved the dependence on the horizon to $T^{2/3}$, with regret
$O((KT)^{2/3}(\log N)^{1/3})$. Banihashem et al.~\cite{bhss} then
improved the dependence on the number of actions, obtaining
$O(T^{2/3}(K\log N)^{1/3})$ regret. More recently, Buzaglo and
Hazan~\cite{bh} obtained $\widetilde O(T^{2/3})$ dependence on the
horizon without requiring auxiliary context samples. Despite these
advances, achieving the optimal $\sqrt{T}$ dependence on the horizon
with an oracle-efficient algorithm has remained an outstanding open
problem.

In this paper, we resolve this problem up to logarithmic factors. We
give an oracle-efficient algorithm with expected regret
\[
  O\!\left(\sqrt{KT\log(2N)}\,\log(2KT)\right).
\]
Following prior work, we consider a hybrid
setting in which contexts are drawn i.i.d.\ and the adversary fixes an
arbitrary sequence of loss functions before play. The learner can draw
independent auxiliary samples from the context distribution.
The learner has access to independent samples
from the context distribution and to an exact offline oracle that
returns the minimum total cost over the policy class. Both the number
and the size of the oracle queries are polynomial in $T$, $K$, and
$\log(2N)$ (Theorem~\ref{main}).

\section{Model and Main Result}\label{sec:model}
Let $T\ge1$, $K\ge2$, and let $\Pi$ be a nonempty finite class of $N$
measurable policies $\pi:\cX\to[K]$. Before play the adversary fixes
measurable losses $c_t:\cX\to[0,1]^K$. The actual contexts $x_t$ are
independent draws from $D$. After observing $x_t$, the learner chooses
$q_t$, draws $A_t\sim q_t$ with fresh randomness, and observes only
$c_t(x_t,A_t)$. Independent auxiliary samples from $D$ are available.
Given a list of contexts and real cost vectors, the value oracle returns
\[
 \mathrm{Val}((y_i,v_i)_i)=\min_{\pi\in\Pi}\sum_i v_i(\pi(y_i)).
\]
We count value-oracle calls, auxiliary context samples, draws from the
specified elementary distributions, and exact real-arithmetic operations.
The learner is given $L=\log(2N)$ or an upper bound $L_0\ge\log(2N)$.
Each logarithm and square root used during initialization counts as one
exact real-arithmetic operation.
\begin{theorem}\label{main}
Let $L=\log(2N)$. Algorithm~\ref{alg}, tuned with $L$, satisfies
\[
 \E\Reg_T:=\E\!\left[\sum_{t=1}^T c_t(x_t,A_t)
       -\min_{\pi\in\Pi}\sum_{t=1}^T c_t(x_t,\pi(x_t))\right]
 \le C\sqrt{KT L}\,\log(2KT)
\]
for a universal constant $C$. Its number of oracle calls, query sizes,
auxiliary context samples, storage and real-arithmetic operations are
polynomial in $T,K,L$. A supplied upper bound $L_0\ge L$ may replace $L$
in both the tuning and the bounds.
\end{theorem}
Section~\ref{ideal} reduces regret to a one-round inequality for the ideal
action distribution. Section~\ref{stability} proves its stability estimate.
Section~\ref{finite} implements the distribution to sufficient accuracy
to add at most one to the regret bound.

\section{Ideal Strategy and Regret Reduction}\label{ideal}
Set
\begin{equation}\label{tuning}
 \gamma=(T+1)^{-1},\quad B=K/\gamma,\quad \ell=\log(eK/\gamma),
 \quad n=\ell^{-1}\sqrt{TL/K},\quad \lambda=4n\ell .
\end{equation}
At the start of round $t$ the score of $\pi$ is $S(\pi)=S_{t-1}(\pi)
=-\sum_{i<t}\hat c_i(\pi(x_i))$, where $S_0=0$ and
\begin{equation}\label{ips}
 \hat c_t=e_{A_t}\,c_t(x_t,A_t)/q_t(A_t).
\end{equation}
A pool consists of $M\sim\Pois(n)$ contexts $X_i\sim D$ and vectors
$Z_i\sim\mu=\Lap(1)^{\otimes K}$, all independent conditional on $M$.
The one-dimensional density is $e^{-|z|}/2$. \pagebreak[0] For a measurable scale
$\sigma:\cX\to[1,B]^K$ define
\begin{align*}
 P_\pi(\sigma)
   &=\sum_{i=1}^M \sigma_{\pi(X_i)}(X_i)Z_{i,\pi(X_i)},\\
 F(S,\sigma)
   &=\E\max_{\pi\in\Pi}\bigl[S(\pi)+P_\pi(\sigma)\bigr],\\
 J_S(\sigma)
   &=F(S,\sigma)+\lambda\E_{X\sim D}\sum_{a=1}^K\log\frac{B}{\sigma_a(X)},
 \qquad R(S)=\inf_\sigma J_S(\sigma).
\end{align*}
Choose a minimizing scale $\sigma$ before observing $x=x_t$
(Lemma~\ref{exist}). With an independent $Z\sim\mu$, let
\begin{equation}\label{win}
 W_x=\argmax_\pi\{S(\pi)+P_\pi(\sigma)+\sigma_{\pi(x)}(x)Z_{\pi(x)}\}.
\end{equation}
A fixed order breaks ties. Conditional on $S,x$ define
\begin{equation}\label{mix}
 p_a(x)=\Pr(W_x(x)=a),\qquad q^*_a(x)=(1-\gamma)p_a(x)+\gamma/K .
\end{equation}
Algorithm~\ref{alg:ideal} gives the ideal strategy. Section~\ref{finite}
implements its scale minimization and action probabilities using a value oracle.
\begin{algorithm}[H]
\caption{Poisson playout with exact action probabilities}\label{alg:ideal}
\begin{pseudocode}
\item Set $S_0(\pi)=0$ for every $\pi\in\Pi$. Use the parameters in \eqref{tuning}.
\item For $t=1,\ldots,T$:
\begin{enumerate}[label=(\alph*),itemsep=2pt]
\item Choose $\sigma\in\argmin_\sigma J_{S_{t-1}}(\sigma)$.
\item Receive $x=x_t$. For each $a\in[K]$, compute
\[
 p_a(x)=\Pr\!\left[
   \left(\argmax_{\pi\in\Pi}
     \bigl\{S_{t-1}(\pi)+P_\pi(\sigma)+\sigma_{\pi(x)}(x)Z_{\pi(x)}\bigr\}
   \right)(x)=a\right].
\]
The probability averages over the auxiliary pool defining $P_\pi(\sigma)$
and the independent vector $Z\sim\mu$, holding $S_{t-1},\sigma,x$ fixed.
\item Set $q_t(a)=(1-\gamma)p_a(x)+\gamma/K$ for each $a$. Draw
$A_t\sim q_t$ and observe $c_t(x,A_t)$.
\item Update the policy scores by
\[
 S_t(\pi)=S_{t-1}(\pi)
 -\frac{c_t(x,A_t)}{q_t(A_t)}\,\ind\{\pi(x)=A_t\},
 \qquad \pi\in\Pi.
\]
\end{enumerate}
\end{pseudocode}
\end{algorithm}
Define for any
distribution $q$ with full support, with $c_a=c_t(x,a)$ and
$f_a(\pi)=\ind\{\pi(x)=a\}$,
\[
 \Phi_{S,x}(q)=\sum_a q_a\left[c_a+
       R(S-(c_a/q_a)f_a)-R(S)\right].
\]
This is the conditional mean of the played loss plus the change in $R$.

\begin{theorem}[Ideal one-round bound]\label{idealbound}
At every score $S$ and for every fixed measurable current loss function,
\[
 \E_{x\sim D}\Phi_{S,x}(q^*(x))\le
 u:=\gamma+6\log(N)/n+3K/B .
\]
\end{theorem}
The only stability estimate needed for this theorem is Lemma~\ref{direct}.
For $q=q^*(x)$ set $b_a=c_a/q_a$ and define
\[
 W_a^-=\argmax_\pi[S(\pi)+P_\pi(\sigma)-b_af_a(\pi)],\qquad
 p_a^-=\Pr(W_a^-(x)=a).
\]
\begin{lemma}[One-round stability]\label{direct}
Fix a round $t$ and an incoming score $S=S_{t-1}$. Let $\sigma$ be the
minimizing scale chosen by the ideal strategy at $S$. There exist
nonnegative functions $D_a:\mathcal X\to[0,\infty)$, $a\in[K]$,
depending on $S$ and $\sigma$, such that
\[
  \E_{x\sim D}\!\left[\sum_{a=1}^K D_a(x)\right]
  \le \frac{\log N}{n}.
\]
For every fixed measurable loss function $c_t:\mathcal X\to[0,1]^K$,
let $p_a^-(x)$ be defined as above with $b_a=c_t(x,a)/q_a^*(x)$.
Then, for $D$-almost every $x$, all $a\in[K]$ satisfy
\[
  p_a(x)-p_a^-(x)\le 6D_a(x)+\frac{3}{B}.
\]
\end{lemma}
\begin{proof}[Proof of Theorem~\ref{idealbound} given Lemma~\ref{direct}]
If $A_t=a$, then $S_t=S-b_af_a$. At the realized $W^-_a$,
\begin{align*}
 \max_\pi[S(\pi)+P_\pi(\sigma)]
 &\ge S(W^-_a)+P_{W^-_a}(\sigma)\\
 &=\max_\pi[S(\pi)+P_\pi(\sigma)-b_af_a(\pi)]+b_af_a(W^-_a).
\end{align*}
Here $W^-_a$ and both maxima use the same pool realization. The same scale is
feasible after the update. Its penalty cancels on subtraction, giving
\begin{align*}
 R(S-b_af_a)-R(S)
 &\le J_{S-b_af_a}(\sigma)-J_S(\sigma)\\
 &= F(S-b_af_a,\sigma)-F(S,\sigma)\\
 &\le-b_a\E f_a(W^-_a)=-b_ap_a^- .
\end{align*}
Use $q_ab_a=c_a$ and then $c_a\in[0,1]$ to obtain
\begin{align*}
\Phi_{S,x}(q)&\le \sum_a q_a(c_a-b_ap_a^-)
 =(q-p)\cdot c+\sum_a c_a(p_a-p_a^-)\\
 &\le\gamma\left(K^{-1}\sum_a c_a-p\cdot c\right)+6\sum_a D_a(x)+3K/B .
\end{align*}
The first term is at most $\gamma$. Averaging over $x$ and applying Lemma~\ref{direct} gives
\[
 \E_x\Phi_{S,x}(q^*(x))
 \le\gamma+6\E_x\sum_a D_a(x)+3K/B
 \le\gamma+6\log(N)/n+3K/B .
\]
\end{proof}
\begin{lemma}[Accumulation of one-round error]\label{tel}
Suppose the actually played vector $q_t$ satisfies
$\E[\Phi_{S_{t-1},x_t}(q_t)\mid\cH_{t-1}]\le u+\xi$ at every round.
Here $\cH_{t-1}$ is the past before observing $x_t$. Then
\[
 \E\Reg_T\le Kn+K\lambda\log B+T(u+\xi).
\]
\end{lemma}
\begin{proof}
At scale $\sigma\equiv1$, $P_\pi(1)\le\sum_{i,a}|Z_{i,a}|$ for every $\pi$,
so $R(0)\le Kn+K\lambda\log B$. At any score $S$, the penalty is nonnegative,
and $\E P_\pi(\sigma)=0$ at fixed $\pi,\sigma$. Hence
$R(S)\ge\max_\pi S(\pi)$. Since the loss functions are fixed before play and the action uses fresh randomness,
\[
 \E[\hat c_t(a)\mid x_{1:T},\cH_{t-1},q_t]
 =q_t(a)c_t(x_t,a)/q_t(a)=c_t(x_t,a).
\]
Consequently,
\begin{align*}
 \E[R(S_T)\mid x_{1:T}]
 &\ge\max_\pi\E[S_T(\pi)\mid x_{1:T}]
 =-\min_\pi\sum_t c_t(x_t,\pi(x_t)).
\end{align*}
Telescoping the potential along the realized scores gives
\begin{align*}
 \E\Reg_T
 &\le\E\left[\sum_t c_t(x_t,A_t)+R(S_T)\right]\\
 &=R(0)+\sum_t\E\Phi_{S_{t-1},x_t}(q_t)
 \le R(0)+T(u+\xi).
\end{align*}
\end{proof}
Put $A=\sqrt{KTL}$. Under \eqref{tuning}, $Kn=A/\ell$,
$K\lambda=4A$, $TL/n=A\ell$, and $T\gamma=TK/B\le1$.
Thus the lemma gives $O(A\log(2KT))+T\xi$. Theorem~\ref{idealbound}
gives $\xi=0$ for ideal play; Theorem~\ref{finitebound} below gives
$\xi=1/T$ for oracle play.
\section{Proof of One-round Stability}\label{stability}
Lemma~\ref{direct} requires two estimates. First, we define $D_a(x)$
and prove
\[
  \E_{x\sim D}\sum_{a=1}^K D_a(x)\le \frac{\log N}{n}.
\]
We then use the optimality of the scale $\sigma$ to prove that, for
$D$-almost every $x$ and each $a\in[K]$,
\[
  p_a(x)-p_a^-(x)\le 6D_a(x)+\frac{3}{B}.
\]

\subsection{Information in the inserted context}

\begin{lemma}[Information carried by one inserted vector]\label{info}
Fix the score $S$ and the scale $\sigma$. For each context $x$, let
$C=(M,X_{1:M})$ be the auxiliary pool locations, let $Z\sim\mu$ be
the independent Laplace vector inserted at $x$, and let $W_x$ be
the winning policy in \eqref{win}.

For each $\pi\in\Pi$, write $w_\pi=\Pr(W_x=\pi\mid C)$. When
$w_\pi>0$, define the conditional distribution $\nu_\pi$ by
\[
  \nu_\pi(A)=\Pr(Z\in A\mid C,\,W_x=\pi)
  \qquad\text{for every measurable }A\subseteq\mathbb R^K.
\]
For each action $a\in[K]$, set
\[
  D_a(x)=\E_C\!\left[
    \sum_{\substack{\pi\in\Pi:\,\pi(x)=a\\ w_\pi>0}}
      w_\pi\,\KL(\nu_\pi\|\mu)
  \right].
\]
Then
\[
  \E_{x\sim D}\!\left[\sum_{a=1}^K D_a(x)\right]
  \le \frac{\log N}{n}.
\]
\end{lemma}

\begin{proof}
For each $m\ge1$, let $(X_1,Z_1),\ldots,(X_m,Z_m)$ be independent
pairs with law $D\otimes\mu$, and define
\[
  W_m=\argmax_{\pi\in\Pi}
  \left\{S(\pi)+
    \sum_{i=1}^m \sigma_{\pi(X_i)}(X_i)Z_{i,\pi(X_i)}\right\},
\]
using the same tie-breaking rule as in \eqref{win}.
Write $\rho_m=\Pr(M=m)=e^{-n}n^m/m!$.

\begin{claim}\label{cl:inserted-information}
\[
  n\,\E_{x\sim D}\!\left[\sum_{a=1}^K D_a(x)\right]
  =
  \sum_{m\ge1}\rho_m
  \sum_{i=1}^m I(Z_i;W_m\mid X_{1:m}).
\]
\end{claim}

\begin{proof}[Proof of Claim~\ref{cl:inserted-information}]
Fix $x$ and the pool locations $C$. Write $I_{x,C}(Z;W_x)$ for
the mutual information under these fixed locations. The inserted
vector $Z$ has law $\mu$. Conditional on $W_x=\pi$, it has law
$\nu_\pi$. Thus
\[
  I_{x,C}(Z;W_x)
  =\sum_{\pi:\,w_\pi>0} w_\pi\,\KL(\nu_\pi\|\mu).
\]
Grouping the policies by their action at $x$ and averaging over $C$
gives
\[
  \E_C[I_{x,C}(Z;W_x)]=\sum_{a=1}^K D_a(x).
\]
Now average over an independent $x\sim D$ and condition on $M=k$.
Since $C=(k,X_{1:k})$ on this event,
\begin{equation}\label{eq:info-condition-count}
  \E_{x\sim D}\!\left[\sum_{a=1}^K D_a(x)\right]
  =\sum_{k\ge0}\rho_k\,
    \E_{x,X_{1:k}}\!\left[I_{x,(k,X_{1:k})}(Z;W_x)\right].
\end{equation}

Fix $k$. Label the inserted pair $(X_{k+1},Z_{k+1})=(x,Z)$.
The resulting $k+1$ pairs are independent with law $D\otimes\mu$.
For every policy $\pi$, its score in the inserted experiment is
\begin{align*}
  S(\pi)
  &+\sum_{i=1}^{k}\sigma_{\pi(X_i)}(X_i)Z_{i,\pi(X_i)}
    +\sigma_{\pi(x)}(x)Z_{\pi(x)}\\
  &=S(\pi)+
    \sum_{i=1}^{k+1}\sigma_{\pi(X_i)}(X_i)Z_{i,\pi(X_i)}.
\end{align*}
Hence $W_x=W_{k+1}$ under this labeling. In particular, at fixed
locations $u_1,\ldots,u_{k+1}$,
\[
  I_{u_{k+1},(k,u_{1:k})}(Z;W_{u_{k+1}})
  =I(Z_{k+1};W_{k+1}\mid X_{1:k+1}=u_{1:k+1}).
\]
By definition, averaging the right-hand side over the locations gives
conditional mutual information. Therefore
\begin{align*}
  \E_{x,X_{1:k}}\!\left[I_{x,(k,X_{1:k})}(Z;W_x)\right]
  &=\int I(Z_{k+1};W_{k+1}\mid X_{1:k+1}=u_{1:k+1})\,
      D^{\otimes(k+1)}(du_{1:k+1})\\
  &=I(Z_{k+1};W_{k+1}\mid X_{1:k+1}).
\end{align*}

Substitute into \eqref{eq:info-condition-count}, multiply by $n$,
and set $m=k+1$. The Poisson probabilities satisfy
$n\rho_{m-1}=m\rho_m$, so
\begin{align*}
  n\,\E_{x\sim D}\!\left[\sum_{a=1}^K D_a(x)\right]
  &=\sum_{k\ge0}n\rho_k\,I(Z_{k+1};W_{k+1}\mid X_{1:k+1})\\
  &=\sum_{m\ge1}n\rho_{m-1}\,I(Z_m;W_m\mid X_{1:m})\\
  &=\sum_{m\ge1}\rho_m\,m\,I(Z_m;W_m\mid X_{1:m}).
\end{align*}

Finally, fix $m$ and $i\in[m]$. Swapping pairs $i$ and $m$ preserves
their joint law. It also preserves $W_m$, since it only reorders the
summands in every policy's score. The swap exchanges $Z_i$ and $Z_m$
and reorders the conditioning vector $X_{1:m}$. This reordering is
invertible, so it preserves the conditioning information. Consequently,
\[
  I(Z_m;W_m\mid X_{1:m})=I(Z_i;W_m\mid X_{1:m}),
  \qquad i\in[m].
\]
Summing over $i$ gives
\[
  m\,I(Z_m;W_m\mid X_{1:m})
  =\sum_{i=1}^m I(Z_i;W_m\mid X_{1:m}),
\]
which proves the claim.
\end{proof}

It remains to bound the inner sum in Claim~\ref{cl:inserted-information}.
Fix $m\ge1$, write $\mathbf X=X_{1:m}$, and let
$Z_{<i}=(Z_1,\ldots,Z_{i-1})$. The noise vectors are independent
conditional on $\mathbf X$, so $I(Z_i;Z_{<i}\mid\mathbf X)=0$.
For each $i$, the chain rule gives
\begin{align*}
  I(Z_i;W_m\mid\mathbf X)
  &\le I(Z_i;Z_{<i},W_m\mid\mathbf X)\\
  &=I(Z_i;Z_{<i}\mid\mathbf X)
    +I(Z_i;W_m\mid\mathbf X,Z_{<i})\\
  &=I(Z_i;W_m\mid\mathbf X,Z_{<i}).
\end{align*}
Summing and applying the chain rule once more yields
\begin{align*}
  \sum_{i=1}^m I(Z_i;W_m\mid\mathbf X)
  &\le \sum_{i=1}^m I(Z_i;W_m\mid\mathbf X,Z_{<i})\\
  &=I(Z_{1:m};W_m\mid\mathbf X)\\
  &\le H(W_m\mid\mathbf X)\le\log N,
\end{align*}
since $W_m$ takes values in the $N$-element class $\Pi$.
Substituting this bound into Claim~\ref{cl:inserted-information} gives
\[
  \E_{x\sim D}\!\left[\sum_{a=1}^K D_a(x)\right]
  \le \frac{1}{n}\sum_{m\ge1}\rho_m\log N
  \le \frac{\log N}{n}.
  \qedhere
\]
\end{proof}

\subsection{The minimizing scale}
For each context $x$, let $g(x)=(\pi(x))_{\pi\in\Pi}$ denote its
prediction signature. Write $s=(s_{g,a})$ for a table with one row per
prediction signature $g$ and one column per action $a\in[K]$.
Each entry $s_{g,a}\in[1,B]$ specifies the scale for action $a$ at
every context with signature $g$. Thus the scale defined by $s$ is
\[
  \sigma_a(x)=s_{g(x),a}.
\]

For each prediction signature $g$, let $\omega_g$ be its probability
under the context distribution $D$, and let $\Gamma_+$ be the set of
signatures with positive probability:
\[
  \omega_g=\Pr_{X\sim D}\!\bigl(g(X)=g\bigr),
  \qquad
  \Gamma_+=\{g:\omega_g>0\}.
\]
\begin{lemma}[Scale minimization on prediction signatures]\label{exist}
There is a minimizing scale of the form $\sigma_a(x)=s^*_{g(x),a}$,
where $g(x)=(\pi(x))_{\pi\in\Pi}$. The table can be chosen measurably in $S$.
\end{lemma}

\begin{proof}
Fix a feasible scale $\sigma$. Define
\[
  s_{g,a}=
  \begin{cases}
    \E[\sigma_a(X)\mid g(X)=g], & g\in\Gamma_+,\\
    1, & g\notin\Gamma_+,
  \end{cases}
  \qquad
  \bar\sigma_a(x)=s_{g(x),a}.
\]
Condition on $\mathcal G=(M,(g(X_i),Z_i)_{i=1}^M)$. The signature
$g(X_i)$ determines $\pi(X_i)$, and the noise is independent of the
contexts. Hence
\[
  \E[P_\pi(\sigma)\mid\mathcal G]
  =\sum_{i=1}^M s_{g(X_i),\pi(X_i)}Z_{i,\pi(X_i)}
  =P_\pi(\bar\sigma).
\]
Conditional Jensen for the maximum and for $u\mapsto\log(B/u)$ gives
\begin{align*}
  \max_\pi[S(\pi)+P_\pi(\bar\sigma)]
  &\le \E\!\left[\max_\pi[S(\pi)+P_\pi(\sigma)]
                  \,\middle|\,\mathcal G\right],\\
  \log(B/s_{g,a})
  &\le \E[\log(B/\sigma_a(X))\mid g(X)=g]
  \qquad(g\in\Gamma_+).
\end{align*}
Take expectations in the first line. In the second, multiply by
$\lambda\omega_g$ and sum over $g,a$. Adding yields
$J_S(\bar\sigma)\le J_S(\sigma)$.

Conversely, every table $s\in[1,B]^{K|\Gamma_+|}$ defines a feasible
scale by setting its entries on zero-mass signatures to one. Thus
\[
  \inf_\sigma J_S(\sigma)
  =\inf_{s\in[1,B]^{K|\Gamma_+|}}J_S(s),
  \qquad
  J_S(s)=F(S,s)+\lambda\sum_{g\in\Gamma_+}\omega_g
                          \sum_{a=1}^K\log(B/s_{g,a}).
\]
The pool term $F(S,s)$ is convex in $s$ as an expected maximum of
affine functions. It is jointly continuous in $(S,s)$ since
\[
  |F(S,s)-F(S',s')|
  \le \|S-S'\|_\infty
     +\|s-s'\|_\infty\,\E\sum_{i=1}^M\sum_{a=1}^K|Z_{i,a}|
  =\|S-S'\|_\infty+Kn\|s-s'\|_\infty.
\]
The penalty is strictly convex on the table because each coordinate
has second derivative $\lambda\omega_g/s_{g,a}^2>0$. Hence $J_S$
attains a unique minimum $s^*(S)$ on the compact box.

Finally, if $S_j\to S$ and a subsequence $s^*(S_{j_k})\to t$, then
for every feasible table $u$, joint continuity gives
\[
  J_S(t)=\lim_k J_{S_{j_k}}(s^*(S_{j_k}))
  \le \lim_k J_{S_{j_k}}(u)=J_S(u).
\]
Uniqueness gives $t=s^*(S)$. By compactness,
$s^*(S_j)\to s^*(S)$, so the minimizing table is continuous,
hence measurable, in $S$.

\end{proof}

Use the scale from Lemma~\ref{exist} in the ideal strategy. For any feasible table $s$,
let $W_x(s)$ denote the argmax in \eqref{win} evaluated at $s$, and write
\[
 \alpha_a(x;s)=\E[Z_a\ind\{W_x(s)(x)=a\}].
\]
For $g(x)=g(y)$, every policy has $\pi(x)=\pi(y)$ and the added
score is identical at $x,y$ for the same noise. Thus $\alpha_a(x;s)=\alpha_a(y;s)$.

\begin{lemma}[Derivative of one scale entry]\label{coord}
For $g(x)=g\in\Gamma_+$, the derivative in entry $(g,a)$ equals
\[
 \frac{\partial J_S(s)}{\partial s_{g,a}}
 =\omega_g\left(n\alpha_a(x;s)-\lambda/s_{g,a}\right).
\]
At the boundary use the derivative of the same formula for $J_S$ extended to positive scales.
In particular at $s=s^*$, writing $\alpha_a(x)=\alpha_a(x;s^*)$,
\begin{equation}
 s^*_{g,a}<B\quad\Longrightarrow\quad \alpha_a(x)\ge\lambda/(n s^*_{g,a}).\label{opt}
\end{equation}
\end{lemma}

\begin{proof}

Let $s'$ be the table obtained by increasing $s_{g,a}$ by $\epsilon$.
In the sum defining $P_\pi(s)$, the $i$th term changes only if
$g(X_i)=g$ and $\pi(X_i)=a$. In that case it changes from
$s_{g,a}Z_{i,a}$ to $(s_{g,a}+\epsilon)Z_{i,a}$. Hence
\[
  [S(\pi)+P_\pi(s')]-[S(\pi)+P_\pi(s)]
  =\epsilon\sum_{i=1}^M
    \mathbf 1\{g(X_i)=g,\ \pi(X_i)=a\}\,Z_{i,a}.
\]
Let $W_m$ be the policy maximizing $S(\pi)+P_\pi(s)$ for a pool of
$m$ samples, with the fixed tie-breaking rule. We claim that
\begin{equation}\label{eq:pool-derivative}
  \frac{\partial F(S,s)}{\partial s_{g,a}}
  =\E\!\left[\sum_{i=1}^M
    \ind\{g(X_i)=g,\ W_M(X_i)=a\}Z_{i,a}\right].
\end{equation}

To see this, fix the pool. Each policy's score is affine in $s_{g,a}$,
with slope
\[
  \sum_{i=1}^M \ind\{g(X_i)=g,\ \pi(X_i)=a\}Z_{i,a}.
\]
If two policies differ at $X_i$, condition on the locations and all
noise coordinates except $Z_{i,\pi(X_i)}$. Their score difference is
a constant plus a nonzero multiple of this continuous variable, so
they tie with probability zero. Since $\Pi$ is finite, almost surely
all tied maximizers agree at every sampled location and therefore
have the same slope. The derivative of the maximum is thus the slope
displayed above with $\pi=W_M$.

Every policy's absolute slope is at most $\sum_{i=1}^M|Z_{i,a}|$.
The same bound holds for the absolute difference quotient of the
maximum, and its expectation is $n$. Dominated convergence therefore
gives \eqref{eq:pool-derivative}.

Expanding the right-hand side of \eqref{eq:pool-derivative} by
conditioning on $M=m$, all $m$ summands have the same expectation
by symmetry. Writing $\rho_m=\Pr(M=m)$, we obtain
\begin{align*}
  \frac{\partial F(S,s)}{\partial s_{g,a}}
  &=\sum_{m\ge1}m\rho_m\,
    \E\!\left[\ind\{g(X_m)=g,\ W_m(X_m)=a\}Z_{m,a}\right]\\
  &=n\sum_{k\ge0}\rho_k\,
    \E\!\left[\ind\{g(X_{k+1})=g,\ W_{k+1}(X_{k+1})=a\}
               Z_{k+1,a}\right]\\
  &=n\E_{X\sim D}\!\left[\ind\{g(X)=g\}\alpha_a(X;s)\right]
   =n\omega_g\alpha_a(x;s).
\end{align*}
The second line uses $m=k+1$ and $m\rho_m=n\rho_{m-1}$.
In the third line, the first $k$ pairs form the pool and the last pair
is the independent inserted pair $(X,Z)$. Averaging over
$k\sim\Pois(n)$ and the noise at fixed $X$ gives
$\alpha_a(X;s)=\E[Z_a\ind\{W_X(s)(X)=a\}\mid X]$.
The last equality uses $\Pr(g(X)=g)=\omega_g$ and
$\alpha_a(X;s)=\alpha_a(x;s)$ whenever $g(X)=g(x)=g$.
Only the term $\lambda\omega_g\log(B/s_{g,a})$ in the penalty depends
on $s_{g,a}$. Adding its derivative to the pool derivative gives
\[
  \frac{\partial J_S(s)}{\partial s_{g,a}}
  =n\omega_g\alpha_a(x;s)-\frac{\lambda\omega_g}{s_{g,a}}.
\]
If $s^*_{g,a}<B$, increasing this entry by a sufficiently small amount
is feasible. Since $s^*$ minimizes $J_S$, the right derivative is
nonnegative. Thus, writing $\alpha_a(x)=\alpha_a(x;s^*)$,
\[
  0\le n\omega_g\alpha_a(x)-\frac{\lambda\omega_g}{s^*_{g,a}}
  \quad\Longrightarrow\quad
  \alpha_a(x)\ge\frac{\lambda}{n s^*_{g,a}},
\]
where we used $n\omega_g>0$.

\end{proof}

\subsection{Comparing the two action probabilities}
For the next two lemmas, write $r_\pi(z)$ for the winning probability when
the inserted vector is fixed at $z$. Explicitly, at $C=(m,x_{1:m})$ set
\begin{align*}
B_\pi(u,z)&=S(\pi)+\sum_{i=1}^m \sigma_{\pi(x_i)}(x_i)u_{i,\pi(x_i)}
                  +\sigma_{\pi(x)}(x)z_{\pi(x)},\\
 r_\pi(z)&=\int\ind\{\pi=\argmax_{\pi'}B_{\pi'}(u,z)\}\,\mu^m(du).
\end{align*}
Here $\mu(dz)=2^{-K}e^{-\|z\|_1}dz$. The law defined in
Lemma~\ref{info} satisfies $w_\pi\nu_\pi(dz)=r_\pi(z)\mu(dz)$, since both
sides equal $\Pr(Z\in dz,W_x=\pi\mid C)$.

\begin{lemma}[Midpoint inequality for a winning policy]\label{midpoint}
Fix $x,C,\pi$. For every pair $z,z'\in\R^K$,
\[r_\pi((z+z')/2)\ge\sqrt{r_\pi(z)r_\pi(z')}.\]
\end{lemma}

\begin{proof}
Fix the context $x$, the pool locations $C=(m,x_{1:m})$, and the
policy $\pi$. Write $u=(u_{i,b})_{i\in[m],\,b\in[K]}$ for the noise
at the pool locations and $z\in\mathbb R^K$ for the noise at $x$.
For each policy $\tau$, its score is
\[
  B_\tau(u,z)
  =S(\tau)+\sum_{i=1}^m
    \sigma_{\tau(x_i)}(x_i)u_{i,\tau(x_i)}
    +\sigma_{\tau(x)}(x)z_{\tau(x)}.
\]
Let $A_\pi$ be the set of pairs $(u,z)$ for which $\pi$ wins.
For each competitor $\tau$, membership in $A_\pi$ requires
\[
  \begin{cases}
    B_\pi(u,z)>B_\tau(u,z),
      &\text{if $\tau$ has priority over $\pi$ in a tie},\\
    B_\pi(u,z)\ge B_\tau(u,z),
      &\text{if $\pi$ has priority over $\tau$ in a tie}.
  \end{cases}
\]
Each score difference $B_\pi-B_\tau$ is affine in $(u,z)$.
Thus each comparison defines a half-space, and their intersection
$A_\pi$ is convex.

Now hold $z$ fixed. The only randomness left is the pool noise $u$,
whose density is $2^{-mK}e^{-\|u\|_1}$. Hence
\[
  r_\pi(z)=\int_{\mathbb R^{mK}} f(u,z)\,du,
  \qquad
  f(u,z)=\ind\{(u,z)\in A_\pi\}\,2^{-mK}e^{-\|u\|_1}.
\]
We verify that $f$ is jointly log-concave. Fix $0<t<1$ and set
\[
  u_t=(1-t)u_0+tu_1,\qquad z_t=(1-t)z_0+tz_1.
\]
If both $(u_0,z_0)$ and $(u_1,z_1)$ belong to $A_\pi$, convexity
gives $(u_t,z_t)\in A_\pi$. Since
$\|u_t\|_1\le(1-t)\|u_0\|_1+t\|u_1\|_1$, we obtain
\[
  f(u_t,z_t)
  =2^{-mK}e^{-\|u_t\|_1}
  \ge f(u_0,z_0)^{1-t}f(u_1,z_1)^t.
\]
If either endpoint lies outside $A_\pi$, the right-hand side is zero,
so the same inequality holds.

Pr\'ekopa's marginalization theorem~\cite{prekopa} now
gives log-concavity of $r_\pi(z)=\int f(u,z)\,du$ in $z$.
Taking $t=1/2$ gives
\[
  r_\pi\!\left(\frac{z_0+z_1}{2}\right)
  \ge \sqrt{r_\pi(z_0)\,r_\pi(z_1)}.
  \qedhere
\]
\end{proof}

\begin{lemma}[Comparison at a fixed penalty]\label{compare}
Fix $x,a$ and put $v=b_a/\sigma_a(x)$. Then
\begin{equation}
 p_a-p_a^-\le 3D_a(x)+\tfrac32p_a\min\{v,1\}^2.\label{comparison}
\end{equation}
\end{lemma}

\begin{proof}

For the $a$th standard basis vector $e_a$,
\[
 \sigma_{\pi(x)}(x)(-ve_a)_{\pi(x)}
 =-v\sigma_a(x)\ind\{\pi(x)=a\}=-b_a\ind\{\pi(x)=a\}.
\]
Consequently the two probabilities being compared are
\begin{align*}
p_a&=\E_C\sum_{\pi(x)=a}\int r_\pi(z)\,\mu(dz)
     =\E_C\sum_{\pi(x)=a}w_\pi,\\
 p_a^-&=\Pr\!\left\{\left(\argmax_\pi[S(\pi)+P_\pi(\sigma)
                    -b_a\ind\{\pi(x)=a\}]\right)(x)=a\right\}
     =\E_C\sum_{\pi(x)=a}r_\pi(-ve_a).
\end{align*}
Write $r=r_\pi$, $w=w_\pi$, $\nu=\nu_\pi$, and $J(z)=-2ve_a-z$. The
midpoint of $z,J(z)$ is $-ve_a$, so Lemma~\ref{midpoint} gives $r(-ve_a)\ge\sqrt{r(z)r(J(z))}$.
For any density $Q$, the reflected density is $(JQ)(z)=Q(J(z))$ since $J^{-1}=J$
and $|\det DJ|=1$. Put $h(P,Q)^2=\int(\sqrt P-\sqrt Q)^2=2-2\int\sqrt{PQ}$.
For $w>0$, use $\nu(z)=r(z)\mu(z)/w$:
\begin{align*}
\int\sqrt{\mu(z)\mu(Jz)}\,dz
 &\le\left(\int\mu(z)\,dz\int\mu(Jz)\,dz\right)^{1/2}=1,\\
 r(-ve_a)&\ge r(-ve_a)\int\sqrt{\mu(z)\mu(Jz)}\,dz\\
 &\ge\int\sqrt{r(z)r(Jz)\mu(z)\mu(Jz)}\,dz\\
 &=w\int\sqrt{\nu(z)(J\nu)(z)}\,dz
 =w\left(1-\tfrac12 h(\nu,J\nu)^2\right).
\end{align*}
When $w=0$, $w-r(-ve_a)\le0$ and this policy can be dropped from the upper bound.
For $w > 0$, the triangle inequality along $\nu,\mu,J\mu,J\nu$ and a change of variables give
\[
 h(\nu,J\nu)\le h(\nu,\mu)+h(\mu,J\mu)+h(J\mu,J\nu),
 \qquad h(J\mu,J\nu)=h(\mu,\nu).
\]
Also
\[
 h(\nu,\mu)^2
 =2\int_{\{\nu>0\}}\nu\left(1-\sqrt{\mu/\nu}\right)
 \le 2\int_{\{\nu>0\}}\nu\left(-\log\sqrt{\mu/\nu}\right)
 =\KL(\nu\|\mu).
\]

Let $\varphi(t)=\tfrac12 e^{-|t|}$ denote the standard Laplace density.
Since $\mu(z)=\prod_{b=1}^K\varphi(z_b)$, $J(z)=-2ve_a-z$, and
$\varphi$ is even, we have
\[
  (J\mu)(z)=\mu(J(z))
  =\varphi(z_a+2v)\prod_{b\ne a}\varphi(z_b).
\]
Put $g_0(t)=\sqrt{\varphi(t)}$. Integrating out the other $K-1$ coordinates and using Cauchy--Schwarz gives
\begin{align*}
h(\mu,J\mu)^2&=\int[g_0(t+2v)-g_0(t)]^2dt\\
 &\le 2v\int_0^{2v}\int |g_0'(t+s)|^2\,dt\,ds
 = (2v)^2/4=v^2 ,
\end{align*}
since $|g_0'|=g_0/2$ almost everywhere and $\int g_0^2=1$. Using $(A+B+C)^2\le3(A^2+B^2+C^2)$ now gives
\begin{align*}
w-r_\pi(-ve_a)
 &\le\tfrac32 w\,[h(\nu,\mu)^2+h(\mu,J\mu)^2+h(J\mu,J\nu)^2]\\
 &\le 3w\KL(\nu\|\mu)+\tfrac32 w v^2,\\
 p_a-p_a^-
 &=\E_C\sum_{\pi(x)=a}(w_\pi-r_\pi(-ve_a))
 \le 3D_a(x)+\tfrac32 p_a v^2 .
\end{align*}
For $v>1$, $p_a-p_a^-\le p_a\le 3D_a(x)+\tfrac32p_a$.
The two ranges therefore give the square of $\min\{v,1\}$ in \eqref{comparison}.

\end{proof}

\begin{lemma}[Cost of the shift]\label{absorb}
At a minimizing scale, for $g(x)\in\Gamma_+$ and $d=\min\{b_a/\sigma_a(x),1\}$,
\[
 p_a d^2\le\begin{cases}2/B,&\sigma_a(x)=B,\\
 \tfrac32D_a(x),&\sigma_a(x)<B.\end{cases}
\]
\end{lemma}

\begin{proof}

Since $\gamma\le1/2$, \eqref{mix} gives $p_a\le2q_a$. Write $v=b_a/\sigma_a(x)$,
so $q_av=c_a/\sigma_a(x)$. When $\sigma_{a}(x)=B$, we have $p_a d^2\le2q_av\le2/B$.
Suppose $\sigma_a(x)<B$. By Lemma~\ref{coord} and the parameter choice
$\lambda=4n\ell$ with $\ell\ge 1$, we have
\[
  \alpha_a(x)\ge \frac{\lambda}{n\sigma_a(x)}
  \ge \frac{4}{\sigma_a(x)}.
\]
Since $q_av=c_a/\sigma_a(x)$ and $c_a\le 1$, it follows that
\[
  \alpha_a(x)\ge \frac{4}{\sigma_a(x)}
  \ge 4q_av \ge 2p_a d,
\]
where the last inequality uses $p_a\le 2q_a$ and $d=\min\{v,1\}\le v$.

It remains to turn the lower bound on $\alpha_a(x)$ into a lower bound
on $D_a(x)$. Set $\theta=d/2$. We will show that
\[
  D_a(x)\ge \theta\alpha_a(x)+p_a\log(1-\theta^2).
\]

Fix the pool locations $C$ and a policy $\pi$ with $w_\pi>0$.
Recall that $\mu$ is the law of the inserted noise vector $Z$, and
$\nu_\pi$ is its conditional law given $C$ and $W_x=\pi$.
To relate $\KL(\nu_\pi\|\mu)$ to $\E_{\nu_\pi}[Z_a]$, define an
auxiliary law $\mu_\theta$ for the same vector by giving each $z$
weight $e^{\theta z_a}$ relative to $\mu$. Dividing by the total
weight gives
\[
  \mu_\theta(dz)=\frac{e^{\theta z_a}}{M_\theta}\,\mu(dz),
  \qquad
  M_\theta=\int e^{\theta z_a}\,\mu(dz).
\]
Since $Z_a$ has density $e^{-|t|}/2$ and $0\le\theta\le1/2$,
\[
  M_\theta
  =\frac12\int_0^\infty e^{-(1-\theta)t}\,dt
   +\frac12\int_0^\infty e^{-(1+\theta)t}\,dt
  =\frac{1}{1-\theta^2}.
\]
By construction, $\log(d\mu_\theta/d\mu)(z)=\theta z_a-\log M_\theta$.
Hence
\begin{align*}
  \KL(\nu_\pi\|\mu)
  &=\KL(\nu_\pi\|\mu_\theta)
    +\E_{\nu_\pi}\!\left[\log\frac{d\mu_\theta}{d\mu}(Z)\right]\\
  &\ge \theta\E_{\nu_\pi}[Z_a]-\log M_\theta.
\end{align*}

Multiply by $w_\pi$, sum over policies with $\pi(x)=a$ and $w_\pi>0$,
and average over $C$. By the definition of $\nu_\pi$,
\begin{align*}
  \E_C\sum_{\substack{\pi(x)=a\\w_\pi>0}}w_\pi\E_{\nu_\pi}[Z_a]
  &=\E[Z_a\ind\{W_x(x)=a\}]=\alpha_a(x),\\
  \E_C\sum_{\pi(x)=a}w_\pi
  &=\Pr(W_x(x)=a)=p_a.
\end{align*}
The weighted sum of the KL terms is $D_a(x)$, so
\[
  D_a(x)\ge\theta\alpha_a(x)-p_a\log M_\theta
          =\theta\alpha_a(x)+p_a\log(1-\theta^2).
\]

Finally, $\theta=d/2$ gives $\theta^2=d^2/4\le1/4$, so
\[
  -\log(1-\theta^2)
  =\int_0^{\theta^2}\frac{dt}{1-t}
  \le\frac{\theta^2}{1-\theta^2}
  \le\frac{d^2}{3}.
\]
Using $\alpha_a(x)\ge 2p_a d$ from above, we conclude that
\[
  D_a(x)\ge\frac d2\,\alpha_a(x)-\frac13p_a d^2
          \ge\frac d2(2p_a d)-\frac13p_a d^2
          =\frac23p_a d^2.
\]
\end{proof}

\begin{proof}[Proof of Lemma~\ref{direct}]
Lemma~\ref{info} gives the expectation bound. For a positive-mass signature, put $d=\min\{b_a/\sigma_a(x),1\}$ and apply
Lemmas~\ref{compare} and~\ref{absorb} to obtain
\[
 p_a-p_a^-\le 3D_a(x)+\tfrac32p_ad^2
 \le\begin{cases}
 (3+\tfrac32\cdot\tfrac32)D_a(x)=\tfrac{21}{4}D_a(x),&\sigma_a(x)<B,\\
 3D_a(x)+\tfrac32\cdot 2/B=3D_a(x)+3/B,&\sigma_a(x)=B .
 \end{cases}
\]

Both bounds are at most $6D_a(x)+3/B$. This proves Lemma~\ref{direct}, since the positive-mass signatures have total probability one.
\end{proof}

\section{Oracle Implementation}\label{finite}
Fix the past $H=((x_i,\hat c_i))_{i<t}$ and its score $S$. The algorithm uses
projected stochastic gradient descent to approximate the minimizer of $J_S$.
Its random iterates are functions of a context. A stored list of draws lets us
evaluate those functions only where they are needed. After $x_t$ arrives,
independent playouts estimate the winning probabilities at that context.

\begin{theorem}[Oracle round]\label{finitebound}
On each round Algorithm~\ref{alg} uses polynomial resources in $T,K,L$ and satisfies
\[
 \E[\Phi_{S,x_t}(q_t)\mid\cH_{t-1}]
 \le \E_{x\sim D}\Phi_{S,x}(q^*(x))+1/T .
\]
\end{theorem}
Together with Theorem~\ref{idealbound} and Lemma~\ref{tel}, this proves
Theorem~\ref{main}. We give the algorithm first and then prove its accuracy and cost.

\subsection{Sampling and evaluation}
Use $n,\lambda,B,\gamma$ from \eqref{tuning}. The following choices fix the
precision, iteration count, number of probability samples, and pool cap:
\begin{align*}
 \delta&=\frac{1}{T(1+K/\gamma)},&
 \varepsilon&=\frac{\lambda\delta^2}{32B^2(n+1)},&
 m&=\lceil16K/\delta^2\rceil,\\
 D_0&=B\sqrt K,& G^2&=K(4n^2+2\lambda^2),&
 J&=\max\{1,\lceil D_0^2G^2/\varepsilon^2\rceil\},\\
 \eta&=D_0/(G\sqrt J),&
 M_{\max}&=\left\lceil(e-1)n+\log(4(J+m)/\delta)\right\rceil .
\end{align*}
Here $D_0$ bounds the diameter of the feasible scales, and $G^2$ bounds
the expected squared norm of one stochastic gradient (Lemma~\ref{lem:opt}).
A pool $U$ consists of pairs $(Y,Z)$ with $Y\sim D$, $Z\sim\mu$. Products
$d\odot Z$ are coordinatewise. All fresh draws in the algorithms are independent.
Write $\mathbf 1$ for the all-ones vector of the required dimension.
\begin{algorithm}[H]
\caption{One bandit round given history $H$}\label{alg}
\small
\begin{pseudocode}
\item Draw $J+m$ capped Poisson counts. If any exceeds $M_{\max}$, receive $x_t$,
set $q=\mathbf1/K$, and continue at line 5. Otherwise use these counts in order.
\item Start with an empty list of training records. For $j=0,\dots,J-1$:
\begin{enumerate}[label=(\alph*),itemsep=1pt]
\item Draw a pool $U_j$ of the specified size. For every $(Y,Z)\in U_j$, call
Algorithm~\ref{replay} at $Y$ on the first $j$ records, obtaining $(d,u)$.
Store $(Y,d\odot Z)$ in the list $P_j$.
\item Draw one $Z^j\sim\mu$ and append $(P_j,Z^j)$ to the records.
\end{enumerate}
\item Receive $x=x_t$. Call Algorithm~\ref{replay} at $x$ on all $J$ records
and set $\bar s(x)=u/J$ from its second output.
\item For $k=1,\dots,m$, draw a fresh pool. At each of its locations $Y$,
evaluate $\bar s(Y)=u/J$ in the same way. Form the list
$P'_k=((Y,\bar s(Y)\odot Z):(Y,Z)\text{ in this pool})$.
Draw $Z'_k\sim\mu$ and set
\[
 A'_k=\routine{Winner}(H,P'_k,x,\bar s(x)\odot Z'_k).
\]
After all $m$ draws, compute for each $a\in[K]$
\[
 \hat p_a=\frac1m\sum_{k=1}^m\ind\{A'_k=a\},
 \qquad q_a=(1-\gamma)\hat p_a+\gamma/K.
\]
\item Draw $A_t\sim q$, observe $c_t(x_t,A_t)$, and append
$(x_t,e_{A_t}c_t(x_t,A_t)/q_{A_t})$ to $H$.
\end{pseudocode}
\end{algorithm}
\begin{algorithm}[H]
\caption{$\routine{Winner}(H,P,y,v)$}\label{winner}
\small\begin{pseudocode}
\item For each $a\in[K]$ obtain $V_a$ by querying $\mathrm{Val}$ on the concatenation
$H,\ ((Y,-u):(Y,u)\in P),\ (y,-v-e_a)$.
\item Return the smallest $a$ attaining $\min_b V_b$.
\end{pseudocode}
\end{algorithm}

\begin{algorithm}[H]
\caption{Evaluate a scale at context $y$ using records $(P_i,Z^i)_{0\le i<j}$ and history $H$}\label{replay}
\small\begin{pseudocode}
\item Initialize $d=\mathbf1$ and $u=0$ in $\R^K$.
\item For $i=0,\ldots,j-1$:
\begin{enumerate}[label=(\alph*),itemsep=1pt]
\item Set $u\gets u+d$ and $A\gets\routine{Winner}(H,P_i,y,d\odot Z^i)$.
\item For every $a\in[K]$ set
\[
 d_a\gets\operatorname{clip}_{[1,B]}
        \big(d_a-\eta[n Z^i_a\ind\{A=a\}-\lambda/d_a]\big).
\]
\end{enumerate}
\item Return $(d,u)$, the current iterate $s^j(y)$ and the sum $\sum_{i<j}s^i(y)$.
\end{pseudocode}
\end{algorithm}

\begin{lemma}[Replay computes the explicit SGD iterates]\label{lem:replay-exact}
Fix the history $H$ and the training draws $(U_j,Z^j)_{0\le j<J}$.
Consider explicit projected SGD starting from $s^0(y)=\mathbf 1$ and
applying the update in Algorithm~\ref{replay} at every context $y$
in every step. For each $0\le j\le J$ and every $y$,
Algorithm~\ref{replay} on the first $j$ records of Algorithm~\ref{alg}
returns
\[
  (d,u)=\left(s^j(y),\,\sum_{i=0}^{j-1}s^i(y)\right).
\]
\end{lemma}

\begin{proof}
We induct on $j$. With no records, replay returns
$(\mathbf 1,0)=(s^0(y),0)$ for every $y$.

Suppose the claim holds after $j$ records. To build record $j$,
Algorithm~\ref{alg} calls replay at each location $Y$ in the new pool
$U_j$. By induction, the returned scale is $s^j(Y)$. Thus the stored
pool is exactly the one explicit SGD constructs:
\[
  P_j=\bigl[(Y,s^j(Y)\odot Z):(Y,Z)\in U_j\bigr].
\]
Now fix any query context $y$. After the first $j$ records, replay has
$d=s^j(y)$. On record $j$, replay and explicit SGD therefore compute
the same action
\[
  A_j(y)=\routine{Winner}(H,P_j,y,s^j(y)\odot Z^j).
\]
For each $a\in[K]$, both update the scale to
\[
  s^{j+1}_a(y)
  =\operatorname{clip}_{[1,B]}\!\left(
    s^j_a(y)-\eta\left[
      nZ_a^j\ind\{A_j(y)=a\}-\frac{\lambda}{s^j_a(y)}
    \right]\right).
\]
Replay also adds the pre-update value $s^j(y)$ to $u$, giving
$u=\sum_{i=0}^{j}s^i(y)$. This proves the claim for $j+1$.
\end{proof}

\begin{lemma}[Oracle access and cost]\label{cost}
Algorithm~\ref{winner} returns the winning action in each queried playout
almost surely. Per round, Algorithm~\ref{alg} makes at most
\[
 O\!\left(K\,[M_{\max}J^2+m(M_{\max}+1)(J+1)]\right)
\]
value calls, each on at most $T+M_{\max}+1$ contexts.
\end{lemma}
\begin{proof}
Write $H_\pi=S(\pi)+\sum_{(Y,u)\in P}u_{\pi(Y)}+v_{\pi(y)}$. Then
\[
 -V_a=\max_\pi[H_\pi+\ind\{\pi(y)=a\}],\quad
 -V_a=1+\max_\pi H_\pi\ \Longleftrightarrow\
 \exists\pi\in\argmax H:\pi(y)=a.
\]
Conditional on the other draws, the highest scores of distinct nonempty
action groups are constants plus independent continuous shifts $s_b(y)Z_b$.
They tie with probability zero. Within a training step this holds jointly
for every prediction signature by a finite union: inductively the input scales
depend on $y$ only through its signature. Thus reusing that step at later
query locations preserves its winning action almost surely.

Building record $j$ takes at most $j M_{\max}$ winner calls. Evaluating
$\bar s(x)$ costs $J$ calls, and each probability trial needs at most
$J M_{\max}+1$. Multiplying by $K$ proves the stated bound.
The storage of scaled training records is $O(KJ(M_{\max}+1))$ in addition to
the history and working lists. At most $(J+m)M_{\max}$ auxiliary contexts
are drawn. Each list construction and coordinate update costs polynomially
many operations in the dimensions shown. A capped count can be sampled with
at most $M_{\max}+1$ independent exponential waiting times of mean one:
stop when their sum exceeds $n$ or $M_{\max}+1$ arrivals occur by $n$.
For $k\le M_{\max}$ the probability of $k$ arrivals is the integral of
$e^{-n}$ over $0<t_1<\cdots<t_k<n$, namely $e^{-n}n^k/k!$.
All displayed parameters and the reciprocals of $\gamma,\lambda,\delta,
\varepsilon$ are polynomially bounded in $T,K,L$ since $L\ge\log2$.
\end{proof}
\subsection{Accuracy of the played distribution}
For the next two lemmas allow full Poisson counts. Write
$\bar s=J^{-1}\sum_{j<J}s^j$ and
$\|h\|_D^2=\E_{Y\sim D}\sum_a h_a(Y)^2$.
At each $y$ the stochastic gradient used in Algorithm~\ref{replay} is
\begin{equation}\label{grad}
 v_a^j(y)=n Z^j_a\ind\{\routine{Winner}(H,P_j,y,s^j(y)\odot Z^j)=a\}
              -\lambda/s_a^j(y).
\end{equation}

\begin{lemma}[Optimization error]\label{lem:opt}
With the parameters of Algorithm~\ref{alg}, if $s^*$ is the scale minimizer at $S$, then
\begin{equation}\label{eq:optdistance}
 \E[J_S(\bar s)-R(S)]\le\varepsilon,\qquad
 \E\norm{\bar s-s^*}_D^2\le 2B^2\varepsilon/\lambda.
\end{equation}
\end{lemma}
\begin{proof}
Fix the incoming score $S$ and use uncapped Poisson pools.
By Lemma~\ref{lem:replay-exact}, replay evaluates the explicit projected
SGD iterates for the same training draws. Thus the scale used by the
algorithm is $\bar s=J^{-1}\sum_{j=0}^{J-1}s^j$.

We first identify the conditional mean of one SGD update. Let $\cH_j$
be the history of the first $j$ training steps, and write
$\E_j[\cdot]=\E[\cdot\mid\cH_j]$. Conditional on $\cH_j$, the current
scale $s^j$ is fixed. At a fixed context $y$, the action used in
\eqref{grad} is
\[
  A_j(y)=\routine{Winner}(H,P_j,y,s^j(y)\odot Z^j).
\]
The pool $P_j$ is a fresh ordinary pool scaled by $s^j$, and $Z^j$ is
the independent noise vector inserted at $y$. Thus $A_j(y)$ is the
action of the inserted winner in the experiment defining $\alpha_a(y;s^j)$.
In particular,
\[
  \E_j\!\left[Z_a^j\ind\{A_j(y)=a\}\right]=\alpha_a(y;s^j).
\]
For each $g\in\Gamma_+$, choose a context $y_g$ with $g(y_g)=g$.
Substituting the preceding identity into \eqref{grad} and applying
Lemma~\ref{coord} gives
\[
  \omega_g\E_j[v_a^j(y_g)]
  =\omega_g\left(n\alpha_a(y_g;s^j)-\frac{\lambda}{s^j_{g,a}}\right)
  =\frac{\partial J_S(s^j)}{\partial s_{g,a}}.
\]
Consequently, with $\langle h,k\rangle_D=\E_{Y\sim D}\sum_a h_a(Y)k_a(Y)$,
convexity of $J_S$ gives
\begin{align*}
  J_S(s^j)-R(S)
  &=J_S(s^j)-J_S(s^*)\\
  &\le \sum_{g\in\Gamma_+}\sum_{a=1}^K
       \frac{\partial J_S(s^j)}{\partial s_{g,a}}
       (s^j_{g,a}-s^*_{g,a})\\
  &=\sum_{g\in\Gamma_+}\omega_g\sum_{a=1}^K
       \E_j[v_a^j(y_g)](s^j_{g,a}-s^*_{g,a})
   =\langle\E_j[v^j],s^j-s^*\rangle_D.
\end{align*}

To bound the squared size of the update, use $s_a^j(y)\ge1$ and
$\E_j[(Z_a^j)^2]=2$. For every $y,a$,
\begin{align*}
  \E_j[(v_a^j(y))^2]
  &\le 2n^2\E_j\!\left[(Z_a^j)^2\ind\{A_j(y)=a\}\right]
       +\frac{2\lambda^2}{(s_a^j(y))^2}\\
  &\le 4n^2+2\lambda^2.
\end{align*}
Hence
\[
  \E_j[\norm{v^j}_D^2]
  =\E_{Y\sim D}\sum_{a=1}^K\E_j[(v_a^j(Y))^2]
  \le K(4n^2+2\lambda^2)=G^2,
\]
where $Y$ is an independent draw from $D$.

Since every coordinate of $s^*$ lies in $[1,B]$, clipping the update
can only decrease its distance to $s^*$. Expanding the square gives
\begin{align*}
  \norm{s^{j+1}-s^*}_D^2
  &\le \norm{s^j-\eta v^j-s^*}_D^2\\
  &=\norm{s^j-s^*}_D^2
    -2\eta\langle v^j,s^j-s^*\rangle_D
    +\eta^2\norm{v^j}_D^2.
\end{align*}
Taking $\E_j$ and using the two bounds above yields
\[
  2\eta\bigl(J_S(s^j)-R(S)\bigr)
  \le \norm{s^j-s^*}_D^2
     -\E_j[\norm{s^{j+1}-s^*}_D^2]+\eta^2G^2.
\]
Taking expectations and summing from $j=0$ to $J-1$ now gives
\begin{align*}
  2\eta\sum_{j=0}^{J-1}\E[J_S(s^j)-R(S)]
  &\le \sum_{j=0}^{J-1}
    \left(\E\norm{s^j-s^*}_D^2-\E\norm{s^{j+1}-s^*}_D^2\right)
    +J\eta^2G^2\\
  &=\norm{s^0-s^*}_D^2-\E\norm{s^J-s^*}_D^2+J\eta^2G^2\\
  &\le D_0^2+J\eta^2G^2.
\end{align*}
By convexity at $\bar s$ and the choices
$\eta=D_0/(G\sqrt J)$ and $J\ge D_0^2G^2/\varepsilon^2$,
\begin{align*}
  \E[J_S(\bar s)-R(S)]
  &\le \frac1J\sum_{j=0}^{J-1}\E[J_S(s^j)-R(S)]\\
  &\le \frac{D_0^2+J\eta^2G^2}{2\eta J}
   =\frac{D_0G}{2\sqrt J}+\frac{D_0G}{2\sqrt J}
   =\frac{D_0G}{\sqrt J}
   \le\varepsilon.
\end{align*}

It remains to bound the distance from $\bar s$ to $s^*$. For
$f(z)=\lambda\log(B/z)$ and $1\le z\le B$, we have
$f''(z)=\lambda/z^2\ge\lambda/B^2$. Thus, for $u,v\in[1,B]$
and $0<t<1$,
\[
  f((1-t)u+tv)
  \le (1-t)f(u)+tf(v)
     -\frac{\lambda}{2B^2}t(1-t)(u-v)^2.
\]
Apply this inequality to $u=s^*_a(Y)$ and $v=s_a(Y)$ for any feasible
scale $s$. Summing over $a$ and averaging over $Y$ gives the squared
distance $\E_Y\sum_a(s_a(Y)-s^*_a(Y))^2=\norm{s-s^*}_D^2$.
Adding the convex pool term and using the optimality of $s^*$ gives
\[
  R(S)
  \le J_S((1-t)s^*+ts)
  \le (1-t)R(S)+tJ_S(s)
     -\frac{\lambda}{2B^2}t(1-t)\norm{s-s^*}_D^2.
\]
In particular,
\begin{align*}
  0
  &\le t\bigl(J_S(s)-R(S)\bigr)
       -\frac{\lambda}{2B^2}t(1-t)\norm{s-s^*}_D^2,\\
  J_S(s)-R(S)
  &\ge \frac{\lambda}{2B^2}(1-t)\norm{s-s^*}_D^2.
\end{align*}
Letting $t\downarrow0$ and applying the result to $s=\bar s$ yields
\[
  \E\norm{\bar s-s^*}_D^2
  \le \frac{2B^2}{\lambda}\E[J_S(\bar s)-R(S)]
  \le \frac{2B^2\varepsilon}{\lambda}.
\]
\end{proof}

Fix the score vector $S$ at the start of the round. For a scale $s$
and a context $x$, let $p_a^s(x)$ be the probability that the
winning policy predicts action $a$ at $x$:
\[
  p_a^s(x)=\Pr\!\left(W_x(s)(x)=a\right).
\]
Here $S$, $s$, and $x$ are fixed. The probability is over a fresh
Poisson pool and an independent noise vector inserted at $x$.
The scale $s$ determines the noise multipliers at every pool location
and at $x$.

Let $s^*$ minimize $J_S$. The ideal strategy plays action $a$ at $x$
with probability
\[
  q_a^*(x)=(1-\gamma)p_a^{s^*}(x)+\frac{\gamma}{K}.
\]
The algorithm instead uses the scale $\bar s$ computed by SGD.
After receiving $x$, it holds $\bar s$ and $x$ fixed and runs $m$
independent trials, each with a fresh pool and a fresh noise vector
at $x$. If $A'_k$ is the winning policy's action at $x$ in trial $k$,
then the fraction of trials won by action $a$ and its resulting play
probability are
\[
  \widehat p_a(x)=\frac1m\sum_{k=1}^m\ind\{A'_k=a\},
  \qquad
  \widehat q_a(x)=(1-\gamma)\widehat p_a(x)+\frac{\gamma}{K}.
\]

\begin{lemma}[Accuracy of the action probabilities before capping]
\label{lem:probapprox}
Suppose the algorithm runs with the pool-size cap disabled. For a
fresh context $x\sim D$, its play probabilities satisfy
\begin{equation}\label{eq:proberror}
  \E\!\left[\sum_{a=1}^K
       \left|\widehat q_a(x)-q_a^*(x)\right|\right]
  \le
  \underbrace{2B\sqrt{\frac{(n+1)\varepsilon}{2\lambda}}}
             _{\text{error from using }\bar s}
  +
  \underbrace{\sqrt{\frac Km}}
             _{\text{error from estimating win probabilities}}.
\end{equation}
The expectation is over the draws used by SGD to compute $\bar s$,
the independent context $x$, and the $m$ trials at that context.
\end{lemma}

\begin{proof}
Fix the incoming score $S$ and use uncapped Poisson pools.

\begin{claim}\label{cl:scale-to-action}
For any two fixed feasible scales $s,t$,
\[
  \E_{x\sim D}\norm{p^s(x)-p^t(x)}_1
  \le \sqrt{n+1}\,\norm{s-t}_D.
\]
\end{claim}

\begin{proof}
Write $X_0=x$ for the inserted context and $X_1,\ldots,X_M$ for the
ordinary pool contexts. Let $Q_s$ be the joint law of $M$, these
locations, and the scaled noise coefficients
$V_{i,a}=s_a(X_i)Z_{i,a}$ for $0\le i\le M$ and $a\in[K]$.
Define $Q_t$ analogously.

We first compare one coefficient. If $V\sim\Lap(u)$, then
$\E|V|=u$, so
\[
  \KL(\Lap(u)\|\Lap(v))
  =\E\!\left[\log\frac{v}{u}
       +|V|\left(\frac1v-\frac1u\right)\right]
  =\log(v/u)+u/v-1
  \le \frac{(u-v)^2}{2}.
\]
For the inequality, the expression as a function of $u$ has value
and derivative zero at $u=v$, and second derivative $1/u^2\le1$
on the feasible interval $[1,B]$.

Under $Q_s$ and $Q_t$, the count and locations have the same law.
Conditional on them, the coefficients are independent. The chain
rule for relative entropy therefore gives
\begin{align*}
  \KL(Q_s\|Q_t)
  &=\E\sum_{i=0}^M\sum_{a=1}^K
    \KL\bigl(\Lap(s_a(X_i))\|\Lap(t_a(X_i))\bigr)\\
  &\le \frac12\E\sum_{i=0}^M\sum_{a=1}^K
    (s_a(X_i)-t_a(X_i))^2\\
  &=\frac12\sum_{k\ge0}\Pr(M=k)(k+1)\norm{s-t}_D^2
   =\frac{n+1}{2}\norm{s-t}_D^2.
\end{align*}
The factor $k+1$ counts the inserted context and the $k$ ordinary
contexts, each drawn from $D$.

Given the locations and coefficients, the same rule under either
scale returns $(x,A)$: it maximizes
$S(\pi)+\sum_{i=0}^M V_{i,\pi(X_i)}$ with the fixed tie-breaking
order and returns the winner's action at $x=X_0$. The context
marginal is $D$ under both scales. Thus, writing $Q_s^{x,A}$ and
$Q_t^{x,A}$ for the two resulting laws, the chain rule and data
processing give
\[
  \E_x\KL(p^s(x)\|p^t(x))
  =\KL(Q_s^{x,A}\|Q_t^{x,A})
  \le\KL(Q_s\|Q_t).
\]
Applying Pinsker's inequality at each $x$ yields
\[
  \E_x\norm{p^s(x)-p^t(x)}_1
  \le \sqrt{\E_x\norm{p^s(x)-p^t(x)}_1^2}
  \le \sqrt{2\E_x\KL(p^s(x)\|p^t(x))}
  \le \sqrt{n+1}\,\norm{s-t}_D.
\]
\end{proof}

\begin{claim}\label{cl:empirical-action}
Conditional on the learned scale $\bar s$ and the context $x$,
the $m$ fresh trials satisfy
\[
  \E\!\left[\norm{\widehat p(x)-p^{\bar s}(x)}_1
       \,\middle|\,\bar s,x\right]\le\sqrt{K/m}.
\]
\end{claim}

\begin{proof}
Write $p_a=p_a^{\bar s}(x)$. For each $a$, the $m$ indicators
$\ind\{A'_k=a\}$ are conditionally independent Bernoulli variables
with mean $p_a$. Their average $\widehat p_a(x)$ therefore satisfies
\[
  \E\!\left[(\widehat p_a(x)-p_a)^2\mid\bar s,x\right]
  =\frac{p_a(1-p_a)}{m}.
\]
By Cauchy--Schwarz and Jensen's inequality,
\begin{align*}
  \E\!\left[\norm{\widehat p(x)-p}_1\mid\bar s,x\right]
  &\le \sqrt{K\sum_{a=1}^K
       \E\!\left[(\widehat p_a(x)-p_a)^2\mid\bar s,x\right]}\\
  &=\sqrt{\frac Km\left(1-\sum_{a=1}^Kp_a^2\right)}
   \le\sqrt{K/m}.
\end{align*}
\end{proof}

Since $x\sim D$ is independent of the training draws, we may apply
Claim~\ref{cl:scale-to-action} conditional on $\bar s$. Averaging
over $\bar s$ and using \eqref{eq:optdistance} gives
\[
  \E\norm{p^{\bar s}(x)-p^{s^*}(x)}_1
  \le \sqrt{(n+1)\E\norm{\bar s-s^*}_D^2}
  \le B\sqrt{\frac{2(n+1)\varepsilon}{\lambda}}.
\]
Finally, $\widehat q(x)-q^*(x)
=(1-\gamma)(\widehat p(x)-p^{s^*}(x))$. Inserting
$p^{\bar s}(x)$ and applying Claim~\ref{cl:empirical-action} yields
\begin{align*}
  \E\norm{\widehat q(x)-q^*(x)}_1
  &\le (1-\gamma)\left(
       \E\norm{\widehat p(x)-p^{\bar s}(x)}_1
       +\E\norm{p^{\bar s}(x)-p^{s^*}(x)}_1\right)\\
  &\le \sqrt{\frac Km}
       +2B\sqrt{\frac{(n+1)\varepsilon}{2\lambda}}.
\end{align*}
\end{proof}

\subsection{Regret at the actual score}
\begin{lemma}[Sensitivity to the played probabilities]\label{lip}
If $q_a,r_a\ge\gamma/K$ for all $a$, then
$|\Phi_{S,x}(q)-\Phi_{S,x}(r)|\le(1+K/\gamma)\|q-r\|_1$.
\end{lemma}
\begin{proof}
Fix $S$ and $x$. For each action $a$, write
$f_a(\pi)=\ind\{\pi(x)=a\}$ and define
\[
  g_a(b)=R(S)-R(S-bf_a),\qquad b\ge0.
\]
Thus $g_a(b)$ is the decrease in the potential when every policy
predicting $a$ at $x$ receives a score penalty of $b$. We will bound
how much the term $q_a g_a(c_a/q_a)$ changes when $q_a$ is replaced
by $r_a$.

We first verify the properties of $g_a$ used in this bound.
Let $s_0,s_1$ be minimizing tables at scores $S_0,S_1$, respectively.
For $0\le t\le1$, put $S_t=(1-t)S_0+tS_1$ and
$s_t=(1-t)s_0+ts_1$. The table $s_t$ is feasible. For every fixed
pool, linearity of $P_\pi$ in the table gives
\begin{align*}
  \max_\pi[S_t(\pi)+P_\pi(s_t)]
  &=\max_\pi\Bigl((1-t)[S_0(\pi)+P_\pi(s_0)]
                   +t[S_1(\pi)+P_\pi(s_1)]\Bigr)\\
  &\le (1-t)\max_\pi[S_0(\pi)+P_\pi(s_0)]
       +t\max_\pi[S_1(\pi)+P_\pi(s_1)].
\end{align*}
Taking expectations and using convexity of the log penalty gives
\[
  R(S_t)\le J_{S_t}(s_t)
  \le (1-t)J_{S_0}(s_0)+tJ_{S_1}(s_1)
  =(1-t)R(S_0)+tR(S_1).
\]
Thus $R$ is convex. Applying this inequality to the scores
$S-b_0f_a$ and $S-b_1f_a$ gives
\begin{align*}
  g_a((1-t)b_0+tb_1)
  &=R(S)-R\bigl((1-t)(S-b_0f_a)+t(S-b_1f_a)\bigr)\\
  &\ge (1-t)g_a(b_0)+t g_a(b_1).
\end{align*}

The potential is also nondecreasing in each score entry, since
increasing a score can only increase the maximum at each fixed table.
Adding the same constant to every score shifts every maximum by that
constant, so $R(S-b\mathbf 1)=R(S)-b$. Since $0\le f_a\le\mathbf 1$,
\[
  R(S)-b=R(S-b\mathbf 1)\le R(S-bf_a)\le R(S).
\]
Hence $0\le g_a(b)\le b$ and $g_a(0)=0$. Moreover, if $b_0\le b_1$,
then $R(S-b_1f_a)\le R(S-b_0f_a)$, so $g_a(b_0)\le g_a(b_1)$.

Now fix one action $a$ and two possible probabilities
$\gamma/K\le s\le t$. Since
\[
  \frac{c_a}{t}
  =\frac{s}{t}\frac{c_a}{s}+\left(1-\frac{s}{t}\right)0,
\]
concavity and $g_a(0)=0$ give
$t\,g_a(c_a/t)\ge s\,g_a(c_a/s)$. For the reverse bound on their
difference, use $c_a/t\le c_a/s$ and monotonicity of $g_a$:
\begin{align*}
  0\le t\,g_a(c_a/t)-s\,g_a(c_a/s)
  &=(t-s)g_a(c_a/t)
    +s\bigl(g_a(c_a/t)-g_a(c_a/s)\bigr)\\
  &\le (t-s)g_a(c_a/t)\\
  &\le (t-s)\frac{c_a}{t}
   \le \frac{K}{\gamma}(t-s).
\end{align*}
The last inequality uses $c_a\le1$ and $t\ge\gamma/K$.
Taking $s=\min\{q_a,r_a\}$ and $t=\max\{q_a,r_a\}$ therefore yields,
for every $a\in[K]$,
\[
  \left|q_a g_a(c_a/q_a)-r_a g_a(c_a/r_a)\right|
  \le \frac{K}{\gamma}|q_a-r_a|.
\]

Finally, expand the difference of the two one-round quantities.
From $\Phi_{S,x}(q)=q\cdot c-\sum_a q_a g_a(c_a/q_a)$, we obtain
\begin{align*}
  \left|\Phi_{S,x}(q)-\Phi_{S,x}(r)\right|
  &=\left|
     \sum_{a=1}^K(q_a-r_a)c_a
     -\sum_{a=1}^K
       \bigl(q_a g_a(c_a/q_a)-r_a g_a(c_a/r_a)\bigr)
     \right|\\
  &\le \sum_{a=1}^K |q_a-r_a|c_a
      +\sum_{a=1}^K
       \left|q_a g_a(c_a/q_a)-r_a g_a(c_a/r_a)\right|\\
  &\le \sum_{a=1}^K |q_a-r_a|
      +\frac{K}{\gamma}\sum_{a=1}^K |q_a-r_a|\\
  &=\left(1+\frac{K}{\gamma}\right)\norm{q-r}_1.
\end{align*}
\end{proof}

\begin{proof}[Proof of Theorem~\ref{finitebound}]
Condition on the actual history $\cH_{t-1}$, so the incoming score $S$
is fixed. Write $\Pr_t$ and $\E_t$ for probability and expectation
conditional on this history. Run the capped and uncapped algorithms
with the same context $x_t$, pool counts, pool samples, and inserted
noise vectors. Let $q_t$ and $\widehat q_t$ be their respective play
probabilities at $x_t$.

Let $M_1,\ldots,M_{J+m}$ be the uncapped pool counts and let
$E=\{\max_i M_i>M_{\max}\}$ be the event that the cap is exceeded.
For each $M_i\sim\operatorname{Poisson}(n)$,
\[
  \E_t[e^{M_i}]
  =e^{-n}\sum_{k=0}^{\infty}\frac{(en)^k}{k!}
  =\exp((e-1)n).
\]
By the union bound and Markov's inequality,
\begin{align*}
  \Pr_t(E)
  &\le \sum_{i=1}^{J+m}\Pr_t(e^{M_i}>e^{M_{\max}})\\
  &\le (J+m)\exp((e-1)n-M_{\max})\\
  &\le (J+m)\exp\!\left(-\log\frac{4(J+m)}{\delta}\right)
   =\frac{\delta}{4}.
\end{align*}

On $E^c$, the two algorithms use exactly the same draws and hence
$q_t=\widehat q_t$. On $E$, both outputs are probability vectors, so
\[
  \norm{q_t-\widehat q_t}_1
  \le \sum_{a=1}^K(q_{t,a}+\widehat q_{t,a})=2.
\]
Thus the expected error introduced by the cap is
\[
  \E_t\norm{q_t-\widehat q_t}_1
  =\E_t\!\left[\ind\{E\}\norm{q_t-\widehat q_t}_1\right]
  \le 2\Pr_t(E)\le\frac{\delta}{2}.
\]

For the uncapped output $\widehat q_t$, Lemma~\ref{lem:probapprox}
applies under the full Poisson law, including draws on which $E$ occurs.
Using $\varepsilon=\lambda\delta^2/[32B^2(n+1)]$ and
$m\ge16K/\delta^2$, we obtain
\begin{align*}
  \E_t\norm{\widehat q_t-q^*(x_t)}_1
  &\le 2B\sqrt{\frac{(n+1)\varepsilon}{2\lambda}}
       +\sqrt{\frac Km}\\
  &\le 2B\sqrt{\frac{\delta^2}{64B^2}}
       +\sqrt{\frac{\delta^2}{16}}
   =\frac{\delta}{2}.
\end{align*}
Inserting $\widehat q_t$ between the actual and ideal play
probabilities therefore gives
\[
  \E_t\norm{q_t-q^*(x_t)}_1
  \le \E_t\norm{q_t-\widehat q_t}_1
     +\E_t\norm{\widehat q_t-q^*(x_t)}_1
  \le \frac{\delta}{2}+\frac{\delta}{2}=\delta.
\]

Every coordinate of $q_t$ and $q^*(x_t)$ is at least $\gamma/K$,
including on overflow, when $q_{t,a}=1/K$. Applying
Lemma~\ref{lip} at the same context $x_t$ gives
\begin{align*}
  \E_t[\Phi_{S,x_t}(q_t)]
  &\le \E_t[\Phi_{S,x_t}(q^*(x_t))]
     +\left(1+\frac K\gamma\right)\E_t\norm{q_t-q^*(x_t)}_1\\
  &\le \E_{x\sim D}[\Phi_{S,x}(q^*(x))]
     +\left(1+\frac K\gamma\right)\delta\\
  &=\E_{x\sim D}[\Phi_{S,x}(q^*(x))]+\frac1T.
\end{align*}
The second line uses that $x_t$ has law $D$ conditional on
$\cH_{t-1}$, and the last line uses
$\delta=1/[T(1+K/\gamma)]$. Lemma~\ref{cost} gives the claimed
per-round resource bound.
\end{proof}

 \section*{Acknowledgments and AI Usage}
This work resulted from extensive interaction between the authors and OpenAI
Astra, with the authors providing suggestions and critically checking Astra's
responses through iterative exchanges. OpenAI Astra was also used in writing
and revising the manuscript through detailed prompts from the authors, with
Gemini Argon assisting in some parts. The authors directed the process and
evaluated the proposed ideas and arguments. Lean files formalizing the main theorem are available on
\href{https://github.com/kbanihashem/bandit-oracle-lean}{https://github.com/kbanihashem/bandit-oracle-lean}.

The authors thank Suho Shin and Max Springer for their collaboration on
earlier work on this problem, which led to the publication of \cite{bhss} and Alex Slivkins for originally
suggesting the problem.

\bibliographystyle{alpha}
\bibliography{refs}

\end{document}